\documentclass[runningheads]{llncs}
\usepackage[T1]{fontenc}
\usepackage{booktabs} 

\usepackage{amsmath,amsfonts,bm}
\usepackage{amsmath}

\def\eqref#1{equation~\ref{#1}}

\def\1{\bm{1}}

\def\ve{{\bm{e}}}

\def\vv{{\bm{v}}}
\def\vw{{\bm{w}}}
\def\vx{{\bm{x}}}
\def\vy{{\bm{y}}}

\def\mW{{\bm{W}}}

\DeclareMathAlphabet{\mathsfit}{\encodingdefault}{\sfdefault}{m}{sl}
\SetMathAlphabet{\mathsfit}{bold}{\encodingdefault}{\sfdefault}{bx}{n}
\newcommand{\tens}[1]{\bm{\mathsfit{#1}}}

\def\tW{{\tens{W}}}

\newcommand{\E}{\mathbb{E}}

\DeclareMathOperator*{\argmin}{arg\,min}

\newtheorem{assumption}{Assumption}
\newtheorem{OpenProblem}{Open Problem}

\usepackage{graphicx}
\begin{document}
\title{Learning distinct features helps, provably\thanks{Preprint}}
%

 \author{Firas Laakom\inst{1}\orcidID{0000-0001-7436-5692} \and
 Jenni Raitoharju\inst{2,3}\orcidID{0000-0003-4631-9298} \and
Alexandros Iosifidis\inst{4}\orcidID{0000-0003-4807-1345} \and \\
Moncef Gabbouj\inst{1}\orcidID{0000-0002-9788-2323}}
\authorrunning{F. Laakom et al.}

\institute{Faculty of Information Technology and Communication Sciences \\
Tampere University,  Finland. \and
Faculty of Information Technology, University of Jyväskylä, Finland \and
Programme for Environmental Information, Finnish Environment Institute,  
Jyväskylä, Finland
\and 
Department of Electrical and Computer Engineering \\  Aarhus University, Denmark }

\maketitle              
\begin{abstract}
We study the diversity of the features learned by a two-layer neural network trained with the least squares loss. We measure the diversity by the average $L_2$-distance between the hidden-layer features and theoretically investigate how learning non-redundant distinct features affects the performance of the network. To do so, we derive novel generalization bounds depending on feature diversity based on Rademacher complexity for such networks. Our analysis proves that more distinct features at the network's units within the hidden layer lead to better generalization. We also show how to extend our results to deeper networks and different losses.
\end{abstract}

\section{Introduction}
Neural networks are a powerful class of non-linear function approximators that have been successfully used to tackle a wide range of problems. They have enabled breakthroughs in many tasks, such as image classification \cite{krizhevsky2012imagenet}, speech recognition \cite{hinton2012deep}, and anomaly detection \cite{golan2018deep}.  
However,  neural networks are often over-parameterized, i.e., have more parameters than the data they are trained on. As a result, they tend to overfit to the training samples and not generalize well on  unseen examples \cite{goodfellow2016deep}. Avoiding overfitting has been extensively studied \cite{neyshabur2018role,nagarajan2019uniform,poggio2017theory,dziugaite2017computing,foret2020sharpness} and  various approaches and strategies have been proposed, such as data augmentation \cite{goodfellow2016deep,zhang2018mixup},  regularization \cite{kukavcka2017regularization,bietti2019kernel,arora2019implicit}, and Dropout \cite{hinton2012improving,lee2019meta,li2016improved}, to close the gap between the empirical loss and the expected loss.

Formally, the output of a neural network consisting of P layers can be defined as follows:
\begin{equation} 
f(\vx;\tW) = \rho^P(\mW^P(\rho^{P-1}( \cdots \rho^2(\mW^2 \rho^1(\mW^1 \vx) ))), 
\end{equation} 
where $\rho^i(.)$ is the element-wise activation function, e.g., \textit{ReLU} or \textit{Sigmoid}, of the $i^{th}$ layer and $\tW= \{\mW^1,\dots,\mW^P\}$ are the weights of the network with the superscript denoting the layer. By defining $\Phi(\cdot)=\rho^{P-1}( \cdots \rho^2(\mW^2 \rho^1(\mW^1 \cdot) )) $, the output of neural network becomes 
\begin{equation} 
f(\vx;\tW) = \rho^P(\mW^P\Phi(\vx)),
\end{equation} 
where $\Phi(\vx)= [\phi_1(\vx),\cdots, \phi_M(\vx)] $ is the $M$-dimensional feature representation of the input $\vx$. This way neural networks can be interpreted as a two-stage process, with the first stage being representation learning, i.e., learning $\Phi(\cdot)$, followed by the final prediction layer. Both parts are jointly optimized.  

Learning a rich and diverse set of features, i.e., the first stage, is critical for achieving top performance \cite{arpit2017closer,laakom2022efficient,cogswell2015reducing}. Studying the different properties of the learned features is an active field of research \cite{deng2021discovering,kornblith2021better,deng2021discovering,du2020few}. For example, \cite{du2020few} showed theoretically that learning a good feature representation can be helpful in few-shot learning. In this paper, we focus on the diversity of the features. This property has been empirically studied in \cite{laakom2023wld,laakom2022reducing,cogswell2015reducing} and has been shown to boost performance and reduce overfitting. However, no theoretical guarantees are provided. In this paper, we close this gap and we conduct a theoretical analysis of feature diversity. In particular, we propose to quantify the diversity of  the feature set $\{ \phi_1(\cdot),\cdots, \phi_M(\cdot)\}$ using the average pairwise $L_2$-distance between their outputs. Formally, given a dataset $\{\vx_i\}_{i=1}^{i=N}$, we have
\begin{equation} \label{div_diff} 
diversity =\frac{1}{N} \sum_{k=1}^N   \frac{1}{2M(M-1)} \sum_{i \neq j}^M  \big( \phi_i(\vx_k) - \phi_j(\vx_k) \big)^2.
\end{equation} 
Intuitively, $diversity$ measures how distinct the learned features are. If the mappings learned by two different units are redundant, then, given the same input, both units would yield similar output. This yields in low $L_2$-distance and as a result a low diversity.
In contrast, if the mapping learned by each unit is distinct, the corresponding average distances to the outputs of the other units within the layer are high. Thus, this yields a high global diversity. 

\begin{figure}[t]
\centering
\includegraphics[width=0.5\linewidth]{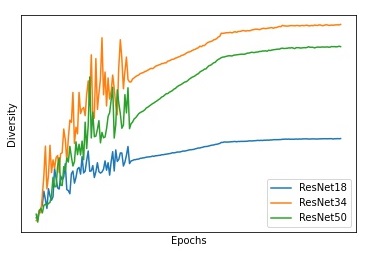}
\caption{Preliminary empirical results for additional motivation to theoretically understand feature diversity. The figure shows diversity versus the number of epochs for three  different ResNet models trained on CIFAR10 dataset.}
\label{cifar_div}
\end{figure}

To confirm this intuition and further motivate the analysis of this attribute, we conduct empirical simulations. We track the diversity of the representation of the last hidden layer, as defined in \eqref{div_diff}, during the training of three different ResNet \cite{he2016deep} models on CIFAR10 \cite{krizhevsky2009learning}. The results are reported in Figure \ref{cifar_div}. Indeed, diversity consistently increases during the training for all the models. This shows that, in order to solve the task at hand, neural networks learn distinct features. \\

\noindent
\textbf{Our contributions:} In this paper, we theoretically investigate diversity in the neural network context and study how learning non-redundant features affects the performance of the model. We derive a bound for the generalization gap which is inversely proportional to the proposed diversity measure showing that learning distinct features helps. In our analysis, we focus on the simple neural network model with one-hidden layer trained with mean squared error. This configuration is simple, however, it has been shown to be convenient and insightful for the theoretical analysis \cite{deng2021adversarial,du2020few,bubeck2021universal}. Moreover, we show how to extend our theoretical analysis to different losses and different network architectures.

Our contributions can be summarized as follows: 
\begin{itemize}
    \item We analyze the effect the feature diversity on the generalization error bound of a neural network. The analysis is presented in Section \ref{sec_theor}. In Theorem \ref{theorm1}, we derive an upper bound for the generalization gap which is inversely proportional to the diversity factor. Thus, we provide theoretical evidence that learning distinct features can help reduce the generalization error.
     \item We extend our analysis to different losses and general multi-layer networks. These results are presented in Theorems \ref{theorm2}, \ref{theorm3}, \ref{theorm_multi}, \ref{theorm4},  and \ref{theorm5}.
\end{itemize}

\noindent
\textbf{Outline of the paper:} The rest of the paper is organized as follows: Section \ref{pre} summarizes the preliminaries for our analysis. Section \ref{sec_theor} presents our main theoretical results along with the proofs. Section \ref{sec_theorext} extends our results for different settings. Section \ref{con} concludes the work with a discussion and several open problems.
\section{PRELIMINARIES} \label{pre}
Generalization theory \cite{shalev2014understanding,kawaguchi2017generalization} focuses on the relation between the empirical loss defined as 
\begin{equation} \label{eq:loss}
\hat{L}(f) = \frac{1}{N} \sum_{i=1}^{N} l\big(f(\vx_i;\tW),y_i \big),
\end{equation}
and the expected risk, for any $f$ in the hypothesis class $\mathcal{F}$, defined as 
\begin{equation}
    L(f) = \E_{(\vx,y)\sim \mathcal{Q}} [l(f(\vx),y)],
\end{equation}
where $\mathcal{Q}$ is the underlying distribution of the dataset and $y_i$ the corresponding label of $x_i$. Let $f^*= \argmin_{f\in \mathcal{F}} L(f)$  be the expected risk minimizer and  $\hat{f}= \argmin_{f\in \mathcal{F}} \hat{L}(f)$ be the empirical risk minimizer. We are interested in the estimation error, i.e.,  $L(f^*) -  L(\hat{f})$, defined as the gap in the loss between both minimizers \cite{barron1994approximation}. The estimation error represents how well an algorithm can learn. It usually depends on the complexity of the hypothesis class and the number of training samples \cite{barron1993universal,zhai2018adaptive}.

Several techniques have been proposed to quantify the generalization error, such as Probably Approximately Correctly (PAC) learning \cite{shalev2014understanding,valiant1984theory}, VC dimension \cite{sontag1998vc}, and the Rademacher complexity \cite{shalev2014understanding}. The Rademacher complexity has been widely used as it usually leads to a tighter generalization error bound than the other metrics \cite{sokolic2016lessons,neyshabur2018role,golowich2018size}. The formal definition of the empirical  Rademacher complexity is given as follows:
\begin{definition} \cite{shalev2014understanding,bartlett2002rademacher}
For a given dataset with N samples $\mathcal{D} = \{\textbf{x}_i,y_i\}_{i=1}^N$ generated by a distribution $\mathcal{Q}$ and for a model space $\mathcal{F} : \mathcal{X} \rightarrow \mathbb{R}$ with a single dimensional output, the empirical Rademacher complexity  $\mathcal{R}_N(\mathcal{F})$ of the set $\mathcal{F}$ is defined as follows:
\begin{equation} 
\mathcal{R}_N(\mathcal{F}) = \E_\sigma \bigg[ \sup_{f \in \mathcal{F} } \frac{1}{N} \sum_{i=1}^{N} \sigma_i f(\vx_i) \bigg],
\end{equation}
where the  variables $\sigma = \{\sigma_1, \cdots, \sigma_N \}$ are independent uniform random  variables  in $\{ -1,1 \}$.
\end{definition}

In this work, we rely on the Rademacher complexity to study diversity. We recall the  following three lemmas related to the Rademacher complexity and the generalization error:
\begin{lemma} \label{complemma} \cite{bartlett2002rademacher} For $\mathcal{F} \in \mathbb{R}^{\mathcal{X}}$, assume that $g:\mathbb{R} \xrightarrow{} \mathbb{R}$ is a $L_{g}$-Lipschitz continuous function and $\mathcal{A} = \{g \circ f : f \in \mathcal{F} \}$. Then we have 
\begin{equation} 
\mathcal{R}_N(\mathcal{A}) \leq L_{g} \mathcal{R}_N(\mathcal{F}).
\end{equation}
\end{lemma}
\begin{lemma} \label{radddd_bound} \cite{xie2015generalization} The Rademacher complexity $\mathcal{R}_N(\mathcal{F})$ of the hypothesis class $\mathcal{F} = \{ f| f(\vx) = \sum_{m=1}^M v_m \phi_m(\vx) = \sum_{m=1}^M v_m \phi(\vw_m^T\vx)  \}$ can be upper-bounded as follows:
\begin{equation} 
\mathcal{R}_N(\mathcal{F} ) \leq \frac{2L_{\rho}C_{134} M}{ \sqrt{N}} + \frac{C_4 |\phi(0)|M}{\sqrt{N}},
\end{equation}
where $C_{134}=C_1C_3C_4$ and $\phi(0)$ is the output of the activation function at the origin. 
\end{lemma}
\begin{lemma} \label{mainlemma} \cite{bartlett2002rademacher}
With a probability of at least $1 - \delta$,  
\begin{equation}
L(\hat{f}) - L(f^*)  \leq  4 \mathcal{R}_N(\mathcal{A}) + B \sqrt{ \frac{2 \log(2/ \delta)}{N}},
\end{equation}
where $B\geq \sup_{\vx,y,f} |l(f(\vx),y)| $ and $\mathcal{R}_N(\mathcal{A})$ is the Rademacher complexity of the loss set $\mathcal{A}$.
\end{lemma}
Lemma \ref{mainlemma} upper-bounds the generalization error using the  Rademacher complexity defined over the loss set and $\sup_{x,y,f} |l(f(x),y)| $. Our analysis aims at expressing this bound in terms of diversity, in order to understand how it affects the generalization. 

In order to study the effect of diversity on the generalization, given a layer with M units $\{ \phi_1(\cdot), \cdots, \phi_M(\cdot) \}$, we make the following assumption:
\begin{assumption}
Given any input $\vx$, we have
\begin{equation} 
\frac{1}{2M(M-1)} \sum_{i \neq j}^M ( \phi_i(\vx) - \phi_j(\vx))^2 \geq d^2_{min}.
\end{equation}
\end{assumption}
$d_{min}$ lower-bounds the average $L_2$-distance between the different units' activations within the same representation layer. Intuitively, if several neuron pairs $i$ and $j$ have similar outputs, the corresponding $L_2$ distance is small. Thus, the lower bound $d_{min}$ is also small and the units within this layer are considered redundant and ``not diverse''. Otherwise, if the average distance between the different pairs is large, their corresponding $d_{min}$ is large and they are considered ``diverse''. By studying how the lower bound $d_{min}$ affects the generalization of the model, we can analyze how the diversity theoretically affects the performance of neural networks. In the rest of the paper, we derive generalization bounds for neural networks using $d_{min}$. 

\section{Learning distinct features helps} \label{sec_theor}

In this section, we derive generalization bounds for neural networks depending on their diversity. Here, we consider a simple tow-layer neural network with a hidden layer composed of $M$ neurons and one-dimensional output trained for a regression task. The full characterization of the setup can be summarized as follows:
\begin{itemize}
    \item The activation function of the hidden layer, $\rho(\cdot)$, is  a positive $L_{\rho}$-Lipschitz continuous function.
    \item The input vector $\vx\in \mathbb{R}^D$ satisfies  $||\vx||_2 \le C_1$ and the output scalar $y\in \mathbb{R}$ satisfies $|y| \le C_2$.

    \item The weight matrix $\mW=[\vw_1,\vw_2,\cdots,\vw_M] \in \mathcal{R}^{D\times M} $ connecting the input to the hidden layer satisfies $||\vw_m||_2 \le C_3$.

    \item The weight vector $\vv \in \mathbb{R}^{M} $ connecting the hidden-layer to the output satisfies $||\vv||_\infty \le C_4$.

    \item The hypothesis class is  $\mathcal{F} = \left\{ f| f(\vx) = \sum_{m=1}^M v_m \phi_m(\vx) = \sum_{m=1}^M v_m \rho(\vw_m^T\vx)  \right\}$.
      
    \item Loss function set is $\mathcal{A}= \left\{ l| l(f(\vx),y) = \frac{1}{2}|f(\vx)-y|^2  \right\}  $.

    \item Given an input $\vx$, $ \frac{1}{2M(M-1)} \sum_{n \neq m}^M ( \phi_n(\vx) - \phi_m(\vx))^2    \geq   d_{min}^2$.
\end{itemize}

Our main goal is to analyze the generalization error bound of the neural network  and to see how its upper-bound is linked to the diversity of the different units, expressed by $d_{min}$. The main result of the paper is presented in Theorem \ref{theorm1}. Our proof consists of three steps: At first, we derive a novel bound for the hypothesis class $\mathcal{F}$ depending on $d_{min}$. Then, we use this bound to  derive bounds for the loss class $\mathcal{A}$ and its Rademacher complexity $\mathcal{R}_N(\mathcal{A})$. Finally, we plug all the derived bounds in Lemma \ref{mainlemma} to complete the proof of Theorem \ref{theorm1}.

The first step of our analysis is presented in  Lemma \ref{supf}:
\begin{lemma} \label{supf}
We have 
\begin{equation} 
\sup_{\vx,f \in \mathcal{F} }  |f(\vx)| \leq  \sqrt{\mathcal{J}},
\end{equation}
where  $ \mathcal{J} = C_4^2 \big( MC^2_5 + M(M-1) (C_5^2 -d_{min}^2 ) \big)$ and $ C_5 =L_{\rho} C_1C_3 +  \phi(0)$,
\end{lemma}
\begin{proof}
\begin{small}
\begin{multline} \label{equ:inequality1}
f^2(\vx) = \left(\sum_{m=1}^M v_m \phi_m(\vx)\right)^2   \leq \left(\sum_{m=1}^M ||\vv||_\infty  \phi_m(\vx)\right)^2  = ||\vv||^2_\infty \left(\sum_{m=1}^M   \phi_m(\vx)\right)^2 \\ \leq C_4^2 \left(\sum_{m=1}^M \phi_m(\vx)\right)^2 
= C_4^2 \left(\sum_{m,n} \phi_m(\vx) \phi_n(\vx)\right) \\ =  C_4^2 \left(  \sum_{m}\phi_m(\vx)^2 +  \sum_{m\neq n} \phi_n(\vx) \phi_m(\vx) \right).
\end{multline}
\end{small}
We have  $\sup_{w,\vx} \phi_m(\vx) = \sup_{w,\vx} \rho(\vw^T\vx)  \leq \sup (L_{\rho}|\vw^T\vx| + \phi(0))  $, because $\rho$ is $L_{\rho}$-Lipschitz. Thus,  $||\phi||_\infty \leq L_{\rho} C_1C_3 +  \phi(0)=C_5 $. For the first term in \eqref{equ:inequality1}, we have  $ \sum_{m}\phi_m(\vx)^2 < M(L_{\rho} C_1C_3 +  \phi(0))^2 =MC^2_5  $.  The second term, using the identity \\ $\phi_m(\vx) \phi_n(\vx) = \frac{1}{2}\left(\phi_m(\vx)^2 + \phi_n(\vx)^2 - (\phi_m(\vx) - \phi_n(\vx))^2 \right)  $, can be rewritten as
\small
\begin{equation}
\sum_{m\neq n} \phi_m(\vx) \phi_n(\vx) = \frac{1}{2}\left( \sum_{m\neq n} \phi_m(\vx)^2 + \phi_n(\vx)^2  -  \Big(\phi_m(\vx) - \phi_n(\vx)\Big)^2 \right).
\end{equation}
In addition, we have  $\frac{1}{2}\sum_{m\neq n} (\phi_m(\vx) - \phi_n(\vx))^2 \geq  M(M-1)d_{min}^2$. Thus, we have:
\begin{equation}
\sum_{m\neq n} \phi_m(\vx) \phi_n(\vx) \leq  \frac{1}{2} \sum_{m\neq n} (2C_5^2) - M(M-1) d_{min}^2 
= M(M-1) (C_5^2 -d_{min}^2) .
\end{equation}
By putting everything back to \eqref{equ:inequality1}, we have:
\begin{equation} 
f^2(\vx) \leq C_4^2 \Big( MC^2_5 + M(M-1) (C_5^2 -d_{min}^2) \Big) = \mathcal{J}.
\end{equation}
Thus, 
$\sup_{\vx,f}  |f(\vx)| \leq \sqrt{\sup_{\vx,f} f(\vx)^2} \leq  \sqrt{\mathcal{J}}. $
\end{proof}
Note that in  Lemma \ref{supf}, we have expressed the upper-bound of $\sup_{\vx,f}  |f(\vx)|$ in terms of $d_{min}$. Using this bound, we can now find an upper-bound for $\sup_{\vx,f,y}  |l(f(\vx),y)|$ in the following lemma:
\begin{lemma} \label{supl}
We have 
\begin{equation} 
\sup_{\vx,y,f} |l(f(\vx),y)|  \leq  \frac{1}{2} (\sqrt{\mathcal{J}} + C_2)^2.
\end{equation}
\end{lemma}
\begin{proof}
We have $\sup_{\vx,y,f} |f(\vx) - y| \leq  \sup_{\vx,y,f} ( |f(\vx)| + |y|) = \sqrt{\mathcal{J}} + C_2$. 
Thus, \\ $\sup_{x,y,f} |l(f(x),y)|  \leq \frac{1}{2} (\sqrt{\mathcal{J}} + C_2)^2 $.
\end{proof}
Next, using the result of lemmas \ref{complemma}, \ref{radddd_bound}, and \ref{supl}, we can derive a bound for the Rademacher complexity of $\mathcal{A}$. We have, thus, expressed all the elements of Lemma  \ref{mainlemma} using the diversity term $d_{min}$. By plugging in the derived bounds in Lemmas \ref{supf}, \ref{supl}, we obtain Theorem \ref{theorm1}.

\begin{theorem} \label{theorm1}
With probability at least $(1 - \delta)$, we have
\begin{equation}
    L(\hat{f}) - L(f^*)  \leq \Big(\sqrt{\mathcal{J}}   + C_2\Big)\frac{A}{\sqrt{N}}
    + \frac{1}{2} ( \sqrt{\mathcal{J}} + C_2)^2 \sqrt{\frac{2 \log(2/ \delta)}{N}},
\end{equation}
where  $C_{134}=C_1C_3C_4$, $\mathcal{J}= C^2_4 \big( MC^2_5 + M(M-1) (C_5^2 -d_{min}^2 ) \big)$, $A=4\Big(2L_{\rho}  C_{134}+ C_4 |\phi(0)| \Big)M$, and   $ C_5 =L_{\rho} C_1C_3 +  \phi(0)$.
\end{theorem}
\begin{proof}
Given that $l(\cdot)$ is $K$-Lipschitz with a constant $K = sup_{\vx,y,f} |f(\vx) - y| \leq  \sqrt{\mathcal{J}} + C_2$,  and using Lemma \ref{complemma}, we can show that  $\mathcal{R}_N(\mathcal{A})\leq K \mathcal{R}_N(\mathcal{F}) \leq  (\sqrt{\mathcal{J}} + C_2) \mathcal{R}_N(\mathcal{F})$.
For $ \mathcal{R}_N(\mathcal{F})$, we use the bound found in Lemma \ref{radddd_bound}. Using Lemmas \ref{mainlemma} and \ref{supl}, we have
\begin{multline}
L(\hat{f}) - L(f^*)  \leq 4 \Big(\sqrt{\mathcal{J}}   + C_2\Big) \Big(2L_{\rho}  C_{134}+ C_4 |\phi(0)| \Big) \frac{M}{\sqrt{N}} +   \frac{1}{2}( \sqrt{\mathcal{J}} + C_2)^2 \sqrt{\frac{2 \log(2/ \delta)}{N}},
\end{multline}
where $C_{134}=C_1C_3C_4$, $\mathcal{J}= C^2_4 \big( MC^2_5 + M(M-1) (C_5^2 -d_{min}^2) \big)$, and   $ C_5 =L_{\rho} C_1C_3 +  \phi(0)$. Thus, setting $A=4\Big(2L_{\rho}  C_{134}+ C_4 |\phi(0)| \Big)M$ completes the proof. 
\end{proof}

Theorem \ref{theorm1} provides an upper-bound for the generalization gap. We note that it is a decreasing function of $d_{min}$. Thus, this suggests that higher  $d_{min}$, i.e., more diverse activations, yields a lower generalization error bound. This shows that learning distinct features helps in neural network context. 

We note that the bound in Theorem \ref{theorm1} is non-vacuous in the sense that it converges to zero when the number of training samples $N$ goes to infinity. Moreover, we note that in this paper we do not claim to reach a tighter generalization bound for neural networks in general \cite{rodriguez2021tighter,jiang2019fantastic,neyshabur2017exploring,dziugaite2017computing}. Our main claim is that we derive a generalization bound which depends on the diversity of learned features, as measured by $d_{min}$. To the best of our knowledge, this is the first work that performs such theoretical analysis based on the average $L_2$-distance between the units within the hidden layer.

\subsection*{Connection to prior studies} 
Theoretical analysis of the properties of the features learned by neural network models is an active field of research.  Feature representation has been theoretically studied in the context of few-shot learning in \cite{du2020few}, where the advantage of learning a good  representation in the case of scarce data was demonstrated. \cite{arora2020provable} showed the same in the context of imitation learning, demonstrating that it  has sample complexity benefits for imitation learning. \cite{wang2021towards} developed similar findings for the self-supervised learning task.
\cite{JMLR_ss} derived novel bounds showing the statistical benefits of multitask representation learning in linear Markov Decision Processes. Opposite to the aforementioned works, the main focus of this paper is not on the large sample complexity problems. Instead, we focused on feature diversity in the learned representation and showed that learning distinct  features leads to better generalization. 

Another line of research related to our work is weight-diversity in neural networks \cite{yu2011diversity,bao2013incoherent,xie2015generalization,xie2017diverse,kwok2012priors}. Diversity in this context is defined based on dissimilarity between the weight component using, e.g., cosine distance and weight matrix covariance \cite{xie2017uncorrelation}.  In \cite{xie2015generalization}, theoretical benefits of weight-diversity have been demonstrated. We note that, in our work, diversity is defined in a fundamentally different way. We do not consider dissimilarity between the parameters of the neural network. Our main scope is the feature representation and, to this end, diversity is defined based on the $L_2$ distance between the feature maps directly and not the weights.  Empirical analysis of the deep representation of neural networks has drawn attention lately \cite{deng2021discovering,kornblith2021better,cogswell2015reducing,laakom2023wld}. For example, \cite{laakom2023wld,cogswell2015reducing} showed empirically that learning decorrelated features reduces overfitting. However, theoretical understanding of the phenomena is lacking. Here, we close this gap by studying how feature diversity affects generalization.


\section{Extensions} \label{sec_theorext}
In this section, we show how to extend our theoretical analysis for classification, for general multi-layer networks, and for different losses.

\subsection{Binary classification}
Here, we extend our analysis of the effect of learning a diverse feature representation on the generalization error to the case of a  binary classification task, i.e., $y \in \{-1,1\} $. Here, we consider the special cases of a hinge loss and a logistic loss. To derive diversity-dependent generalization bounds for these cases, similar to the proofs of Lemmas 7 and 8 in \cite{xie2015generalization}, we can show the following two lemmas:
\begin{lemma} \label{lemmaa2}
Using the hinge loss, we have with probability at least $(1 - \delta)$
\begin{multline} 
L(\hat{f}) - L(f^*)  \leq  4 \Big(2L_{\rho}C_{134} + C_4 |\phi(0)| \Big) \frac{M}{\sqrt{N}}  +  (1+\sqrt{\mathcal{J}}) \sqrt{\frac{2 \log(2/ \delta)}{N}},
\end{multline}
where $C_{134}=C_1C_3C_4$, $\mathcal{J}= C^2_4 ( MC^2_5 + M(M-1) (C_5^2 -d_{min}^2 ) \big)$, and   $ C_5 =L_{\rho} C_1C_3 +  \phi(0)$.
\end{lemma}

\begin{lemma} \label{lemmaa3}
Using the logistic loss $l(f(x),y) = \log(1 + \ve^{-yf(x)})$, we have with probability at least $(1 - \delta)$
\begin{multline} 
L(\hat{f}) - L(f^*)  \leq  \frac{4}{1 + \ve^{\sqrt{-\mathcal{J}}}} \Big(2L_{\rho}C_{134} + C_4 |\phi(0)|\Big) \frac{M}{\sqrt{N}}
+   \log(1+\ve^{\sqrt{\mathcal{J}}}) \sqrt{\frac{2 \log(2/ \delta)}{N}},
\end{multline}
where $C_{134}=C_1C_3C_4$, $\mathcal{J}= C^2_4 ( MC^2_5 + M(M-1) (C_5^2 -d_{min}^2 ) \big)$, and   $ C_5 =L_{\rho} C_1C_3 +  \phi(0)$. \end{lemma}
Using the above lemmas, we can now derive a diversity-dependant bound for the binary classification case. The extensions of Theorem \ref{theorm1} in the cases of a hinge loss and a logistic loss are presented in Theorems \ref{theorm2} and \ref{theorm3}, respectively.

\begin{theorem} \label{theorm2}
Using the hinge loss, with probability at least $(1 - \delta)$, we have
\begin{equation} 
L(\hat{f}) - L(f^*)  \leq  A/\sqrt{N} +  (1+\sqrt{\mathcal{J}}) \sqrt{\frac{2 \log(2/ \delta)}{N}},
\end{equation}
where  $\mathcal{J}= C^2_4 ( MC^2_5 + M(M-1) (C_5^2 -d_{min}^2 ) \big)$, $A=4 \Big(2L_{\rho}C_{134} + C_4 |\phi(0)| \Big)M$, and  $ C_5 =L_{\rho} C_1C_3 +  \phi(0)$.
\end{theorem}
\begin{theorem} \label{theorm3}
Using the logistic loss $l(f(x),y) = \log(1 + \ve^{-yf(x)})$, with probability at least $(1 - \delta)$, we have
\begin{equation} 
L(\hat{f}) - L(f^*)  \leq  \frac{A }{(1 + \ve^{\sqrt{-\mathcal{J}}})\sqrt{N}} +   \log(1+\ve^{\sqrt{\mathcal{J}}}) \sqrt{\frac{2 \log(2/ \delta)}{N}},
\end{equation}
where $\mathcal{J}= C^2_4 ( MC^2_5 + M(M-1) (C_5^2 -d_{min}^2 ) \big)$, $A=4 \Big(2L_{\rho}C_{134} + C_4 |\phi(0)| \Big)M$,  and   $ C_5 =L_{\rho} C_1C_3 +  \phi(0)$.
\end{theorem}
As we can see, also for the binary classification task, the generalization bounds for the hinge and logistic losses are decreasing with respect to $d_{min}$. Thus, this shows that learning distinct features helps and can improve the generalization also in binary classification.

\subsection{Multi-layer networks}
Here, we extend our result for  networks with P ($>1$) hidden layers. We assume that the pair-wise distances between the activations within  layer $p$ are lower-bounded by $d_{min}^{(p)}$. In this case, the hypothesis class can be defined recursively. In addition, we assume that: $||\mW^{(p)}||_{\infty} \leq C^{(p)}_3$ for every $\mW^{(p)}$, i.e., the weight matrix of the $p$-th layer. In this case, the main theorem is extended as follows:
\begin{theorem} \label{theorm_multi}
With  probability of at least $ (1 - \delta)$, we have
\begin{equation}
L(\hat{f}) - L(f^*)  \leq (\sqrt{\mathcal{J}^P}   + C_2) \frac{A}{\sqrt{N}}
+ \frac{1}{2}\left( \sqrt{\mathcal{J}^P} + C_2 \right)^2 \sqrt{\frac{2 \log(2/ \delta)}{N}},
\end{equation}
where $A= 4( (2L_{\rho})^P C_1 C^0_3 \prod^{P-1}_{p=0} \sqrt{M^{(p)}} C_3^{(p)} + |\phi(0)|\sum_{p=0}^{P-1} (2L_{\rho})^{P-1-p} \prod^{P-1}_{j=p} \sqrt{M^j} C_3^j ) $, and $\mathcal{J}^P$ is defined recursively using the following identities: $ \mathcal{J}^0 = C_3^{0} C_1 $ and \\ $\mathcal{J}^{(p)}= M^{(p)} {C^{p}}^2 \big( M^{p2} (L_{\rho}  \mathcal{J}^{p-1}  + \phi(0))^2 - M(M-1) {d_{min}^{(p)}}^2 ) \big)$, for $p=1, \dots,P$.
\end{theorem}
\begin{proof}
Lemma 5 in \cite{xie2015generalization} provides an upper-bound for the hypothesis class. We denote by $\vv^{(p)}$ the outputs of the $p^{th}$ hidden layer before applying the activation function:
\begin{equation}
\vv^0 = [\vw_1^{0^T}\vx, .... , \vw^{0^T}_{M^0}\vx ],
\end{equation}
\begin{equation}
\vv^{(p)} = \left[\sum_{j=1}^{M^{p-1}} w^{(p)}_{j,1} \phi(\vv^{p-1}_j), .... , \sum_{j=1}^{M^{p-1}} w^{(p)}_{j,M^{(p)}} \phi(v^{p-1}_j)  \right],
\end{equation}

\begin{equation}
\vv^{(p)} = \left[{\vw_1^{(p)}}^T \boldsymbol{\phi}^{(p)}, ..., {\vw^{(p)}_{M^{(p)}}}^T \boldsymbol{\phi}^{(p)} \right],
\end{equation}
where $\boldsymbol{\phi}^{(p)}= [ \phi(v^{p-1}_1), \cdots,  \phi(v^{p-1}_{M^{p-1}})  ]$. We have 
$||\vv^{(p)}||^2_2 = \sum_{m=1}^{M^{(p)}} ({\vw^{(p)}_m}^T \boldsymbol{\phi}^{(p)})^2 $
and 
${\vw^{(p)}_m}^T \boldsymbol{\phi}^{(p)} \leq C_3^{(p)} \sum_n \phi^{(p)}_n $. Thus, 
\begin{equation}
||\vv^{(p)}||^2_2 \leq \sum_{m=1}^{M^{(p)}} \left( C_3^{(p)} \sum_n \phi^{(p)}_n \right)^2 = M^{(p)} {C_3^{p}}^2  \left(\sum_n \phi^{(p)}_n \right)^2 
= M^{(p)} {C_3^{p}}^2  \sum_{mn} \phi^{(p)}_m \phi^{(p)}_n.
\end{equation}
We use the same decomposition trick of $\phi^{(p)}_m \phi^{(p)}_n$ as in the proof of Lemma 2. We need to bound $\sup_x \phi^{(p)} $: \\
\begin{equation}
\sup_x \phi^{(p)}< \sup(L_{\rho} |\vv^{p-1}|+ \phi(0)) 
< L_{\rho} ||\vv^{p-1}||^2_2  + \phi(0).
\end{equation}
Thus, we have
\begin{equation}
||\vv^{(p)}||^2_2 \leq  M^{(p)} {C_3^{p}}^2 \big( M^2 (L_{\rho}  ||\vv^{p-1}||^2_2  + \phi(0))^2 
- M(M-1) d_{min}^2 ) \big)  = \mathcal{J}^P.
\end{equation}
We found a recursive bound for  $||\vv^{(p)}||^2_2$ and we note that for $p=0$ we have 
$||\vv^0||^2_2 \leq ||W^0||_{\infty} C_1 \leq C^0_3 C_1 = \mathcal{J}^0 $. Thus, 
\begin{equation}
\sup_{\vx,f^P \in \mathcal{F}^P}  |f(\vx)| = \sup_{\vx,f^P \in \mathcal{F}^P}  |\vv^P| \leq \sqrt{\mathcal{J}^P}.
\end{equation}
By replacing the variables in Lemma \ref{mainlemma}, we have
\begin{multline}
L(\hat{f}) - L(f^*)  \leq 4(\sqrt{\mathcal{J}^P}   + C_2) \Bigg( \frac{(2L_{\rho})^P C_1 C^0_3}{\sqrt{N}} \prod^{P-1}_{p=0} \sqrt{M^{(p)}} C_3^{(p)} \nonumber \\
+ \frac{ |\phi(0)|}{\sqrt{N}}\sum_{p=0}^{P-1} (2L_{\rho})^{P-1-p} \prod^{P-1}_{j=p} \sqrt{M^j} C_3^j \Bigg)
+ \frac{1}{2}\left( \sqrt{\mathcal{J}^P} + C_2 \right)^2 \sqrt{\frac{2 \log(2/ \delta)}{N}},
\end{multline}
Taking $A= 4\big( (2L_{\rho})^P C_1 C^0_3 \prod^{P-1}_{p=0} \sqrt{M^{(p)}} C_3^{(p)} + |\phi(0)|\sum_{p=0}^{P-1} (2L_{\rho})^{P-1-p} \prod^{P-1}_{j=p} \sqrt{M^j} C_3^j \big) $ completes the proof.
\end{proof}
In Theorem \ref{theorm_multi}, we see that $\mathcal{J}^P$ is decreasing with respect to $d^{(p)}_{min}$. This extends our results to the multi-layer neural network case. 
\subsection{ Multiple outputs}
Finally, we consider the case of a neural network with a multi-dimensional output, i.e., $\vy \in R^D$.  In this case, we can extend Theorem \ref{theorm1} with the following two theorems:
\begin{theorem} \label{theorm4}
For a multivariate regression trained with the squared error, there exists a constant A such that, with probability at least $(1 - \delta)$, we have
\begin{equation}
L(\hat{f}) - L(f^*)  \leq (\sqrt{\mathcal{J}} + C_2)\frac{A}{\sqrt{N}} + \frac{D}{2}( \sqrt{\mathcal{J}} + C_2)^2 \sqrt{\frac{2 \log(2/ \delta)}{N}}
\end{equation}
where $\mathcal{J}= C^2_4 ( MC^2_5 + M(M-1) (C_5^2 -d_{min}^2 ) \big)$,   $ C_5 =L_{\rho} C_1C_3 +  \phi(0)$, and  $A= 4D  \Big(2L_{\rho}  C_{134}+ C_4 |\phi(0)|\Big)M$.
\end{theorem}
\begin{proof}
The squared loss $ \frac{1}{2}||f(\vx) - \vy||_2^2 $ can be decomposed into D terms $\frac{1}{2} (f(\vx)_k - y_k)^2$. Using Theorem \ref{theorm1}, we can derive the bound for each term and, thus, we have:
\begin{equation}
L(\hat{f}) - L(f^*)  \leq 4D(\sqrt{\mathcal{J}} + C_2) \Big(2L_{\rho}  C_{134}+ C_4 |\phi(0)|\Big) \frac{M}{\sqrt{N}} + \frac{D}{2}( \sqrt{\mathcal{J}} + C_2)^2 \sqrt{\frac{2 \log(2/ \delta)}{N}},
\end{equation}
where $C_{134}=C_1C_3C_4$, $\mathcal{J}= C^2_4 ( MC^2_5 + M(M-1) (C_5^2 -d_{min}^2 ) \big)$, and   $ C_5 =L_{\rho} C_1C_3 +  \phi(0)$. Taking $A= 4D  \Big(2L_{\rho}  C_{134}+ C_4 |\phi(0)|\Big)M$ completes the proof.
\end{proof} 
\begin{theorem} \label{theorm5}
For a multi-class classification task using the cross-entropy loss, there exists a constant A such that, with probability at least $(1 - \delta)$, we have 
\begin{multline}
L(\hat{f}) - L(f^*)  \leq \frac{A}{(D-1+\ve^{-2\sqrt{\mathcal{J}}})\sqrt{N}} +   \log\Big(1 + (D-1) \ve^{2\sqrt{\mathcal{J}}}\Big)   \sqrt{\frac{2 \log(2/ \delta)}{N}}
\end{multline}
where $\mathcal{J}= C^2_4 ( MC^2_5 + M(M-1) (C_5^2 -d_{min}^2 ) \big)$ and   $ C_5 =L_{\rho} C_1C_3 +  \phi(0)$, and  $A=4D(D-1)\left(2L_{\rho}C_{134} + C_4 |\phi(0)|\right)M $.
\end{theorem}
\begin{proof}
Using Lemma 9 in \cite{xie2015generalization}, we have $\sup_{f,\vx,\vy} l=  \log\big(1 + (D-1) \ve^{2\sqrt{\mathcal{J}}}\big)  $ and $l$ is $\frac{D-1}{D-1+\ve^{-2\sqrt{\mathcal{J}}}}$-Lipschitz. Thus, using the decomposition property of the Rademacher complexity, we have 
\small
\begin{equation}
\mathcal{R}_n(\mathcal{A}) \leq \frac{4D(D-1)}{D-1+\ve^{-2\sqrt{\mathcal{J}}}}  \left(2L_{\rho}C_{134} + C_4 |\phi(0)|\right)\frac{M}{\sqrt{N}}.
\end{equation}
Taking $A=4D(D-1)\left(2L_{\rho}C_{134} + C_4 |\phi(0)|\right)M $ completes the proof.
\end{proof}
Theorems \ref{theorm4} and \ref{theorm5} extend our result for the multi-dimensional regression and classification tasks, respectively. Both bounds are inversely proportional to the diversity factor $d_{min}$. We note that for the classification task the upper-bound is exponentially decreasing with respect to $d_{min}$. This shows that learning a diverse  and rich feature representation yields a tighter generalization gap and, thus, theoretically guarantees a stronger generalization performance.

\section{Discussion and open problems} \label{con}
In this paper, we showed how the diversity of the features learned by a two-layer neural network trained with the least-squares loss affects generalization. We quantified the diversity by the average $L_2$-distance between the hidden-layer features and we derived novel diversity-dependant generalization bounds based on Rademacher complexity for such models. The derived bounds are inversely-proportional to the diversity term, thus demonstrating that more distinct features within the hidden layer can lead to better generalization. We also showed how to extend our results to deeper networks and different losses. 

\begin{figure}[t]
\centering
\includegraphics[width=0.33\linewidth]{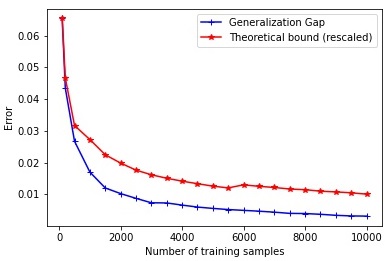}
\includegraphics[width=0.33\linewidth]{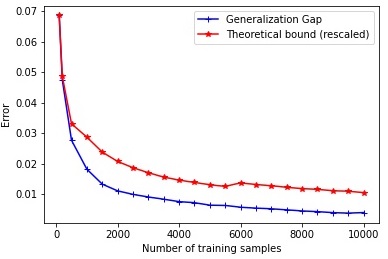}
\includegraphics[width=0.32\linewidth]{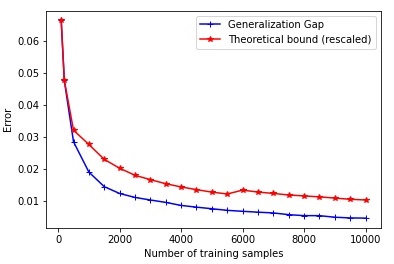}

\caption{Generalization gap, i.e., train error - test error, and the theoretical bound, i.e., $ (C_5^2 - d^2_{min})/\sqrt{N}$, as a function of the number of training samples on MNIST dataset for neural networks with intermediate layer sizes from left to right: 128 (correlation=0.9948), 256 (correlation=0.9939), and 512 (correlation=0.9953). The theoretical term has been scaled in the same range as the generalization gap. All results are averaged over 5 random seeds. }
\label{fig_theorydiv}
\end{figure}

The bound found in Theorem \ref{theorm1} suggests that the generalization gap, with respect to diversity, is inversely proportional to $d_{min}$ and scales as $\sim (C_5^2 - d^2_{min})/\sqrt{N}$. We validate this finding empirically in Figure \ref{fig_theorydiv}. We train a two-layer neural network on the MNIST dataset for 100 epochs using SGD with a learning rate of $0.1$ and batch size of 256. We show the generalization gap, i.e., test error - train error, and the theoretical bound, i.e., $ (C_5^2 - d^2_{min})/\sqrt{N}$, for different training set sizes. $d_{min}$ is the lower bound of diversity. Empirically, it can be estimated as the minimum feature diversity over the training data $S$: $ d_{min}= \min_{x \in S} \frac{1}{2M(M-1)} \sum_{n \neq m}^M ( \phi_n(\vx) - \phi_m(\vx))^2   $. We experiment with different sizes of the hidden layer, namely 128, 256, and 512. The average results using 5 random seeds are reported  for different training sizes in Figure \ref{fig_theorydiv} showing that the theoretical bound  correlates consistently well (correlation $>$ 0.9939) with the generalization error. 

\begin{figure}[t]
\centering
\includegraphics[width=0.33\linewidth]{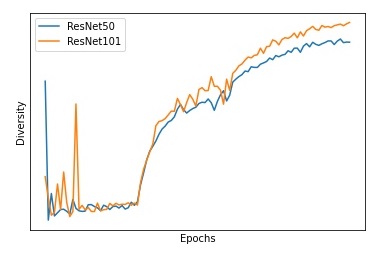}
\includegraphics[width=0.33\linewidth]{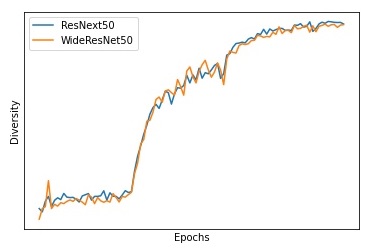}
\includegraphics[width=0.32\linewidth]{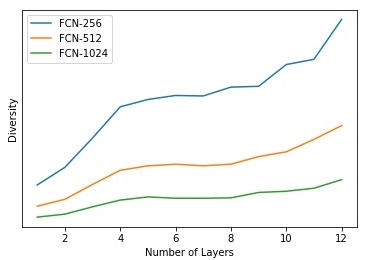}
\caption{From left to right: (a)-(b) Tracking the diversity during the training for different models on ImageNet. (c) Final diversity as a function of depth for different models on MNIST.  }
\label{training_div}
\end{figure}

As shown in Figure \ref{cifar_div}, diversity increases for neural networks along the training phase. To further investigate this observation, we conduct additional experiments on ImageNet \cite{russakovsky2015imagenet} dataset using 4 different state-of-the-art models:  \textbf{ResNet50} and  \textbf{ResNet101}, i.e., the standard ResNet model \cite{he2016deep} with 50 layers and 101 layers,   \textbf{ResNext50}  \cite{xie2017aggregated}, and \textbf{WideResNet50} \cite{BMVC2016_87} with 50 layers. All models are trained with SGD using standard training protocol \cite{zhang2018mixup,huang2017densely,cogswell2015reducing}. We track the diversity, as defined in \eqref{div_diff},  of the features of the last intermediate layer. The results are shown in Figure \ref{training_div} (a) and (b). As it can be seen, SGD without any 
explicit regularization implicitly optimizes diversity and converges toward regions with high features' distinctness. These observations suggest the following conjecture: 
\begin{conjecture} \label{implicit_div}
Standard training with SGD implicitly optimizes the diversity of intermediate features.
\end{conjecture}
Studying the fundamental properties of SGD is extremely important to understand generalization in deep learning \cite{kawaguchi2019gradient,kalimeris2019sgd,volhejn2020does,zou2021benefits,pmlr-v99-ji19a}. Conjecture \ref{implicit_div} suggests a new implicit bias for SGD, showing that it favors regions with high feature diversity.

Another research question related to diversity that is worth investigating is: \textit{How does the network depth affect diversity?}  In order to answer this question, we conduct an empirical experiment using MNIST dataset \cite{lecun1998gradient}. We use fully connected networks (FCNs) with ReLU activation and different depths (1 to 12). We experiment with three models with different widths, namely FCN-256, FCN-512, and FCN-1024, with 256, 512, and 1024 units per layer, respectively. We measure the final diversity of the last hidden layer for the different depths. The average results using 5 random seeds are reported in Figure \ref{training_div} (c). Interestingly, in this experiment, increasing the depth consistently leads to learning more distinct features and higher diversity for the different models. However, by looking at Figure \ref{cifar_div}, we can see that having more parameters does not always lead to higher diversity. This suggests the following open question:
\begin{OpenProblem} \label{deep_div}
When does having more parameters/depth lead to higher diversity? 
\end{OpenProblem}
Understanding the difference between shallow and deep models and why deeper models generalize better is one of the puzzles of deep learning \cite{liao2019generalization,kawaguchi2019depth,poggio2017theory}. The insights gained by studying Open Problem \ref{deep_div} can lead to a novel key advantage of depth: deeper models are able to learn a richer and more diverse set of features.

Another interesting line of research is adversarial robustness 
\cite{NEURIPS2019_36ab6265,wu2021wider,liao2019generalization,mao2020multitask}. Intuitively, learning distinct features can lead to a richer representation and, thus, more robust networks. However, the theoretical link is missing. This leads to the following open problem:
\begin{OpenProblem} \label{prob_corr}
Can the theoretical tools proposed in this paper be used to prove the benefits of feature diversity for adversarial robustness?
\end{OpenProblem}


\subsubsection{Acknowledgements} This work has been supported by the NSF-Business Finland Center for Visual and Decision Informatics (CVDI) project AMALIA. The work of Jenni Raitoharju was funded by the Academy of Finland (project 324475). 
Alexandros Iosifidis acknowledges funding from the European Union’s Horizon 2020 research and innovation programme under grant agreement No 957337.

%
%
%
\bibliographystyle{splncs04}
\bibliography{main}

\begin{thebibliography}{10}
\providecommand{\url}[1]{\texttt{#1}}
\providecommand{\urlprefix}{URL }
\providecommand{\doi}[1]{https://doi.org/#1}

\bibitem{arora2019implicit}
Arora, S., Cohen, N., Hu, W., Luo, Y.: Implicit regularization in deep matrix
  factorization. In: Advances in Neural Information Processing Systems. pp.
  7413--7424 (2019)

\bibitem{arora2020provable}
Arora, S., Du, S., Kakade, S., Luo, Y., Saunshi, N.: Provable representation
  learning for imitation learning via bi-level optimization. In: International
  Conference on Machine Learning. PMLR (2020)

\bibitem{arpit2017closer}
Arpit, D., Jastrzkebski, S., Ballas, N., Krueger, D., Bengio, E., Kanwal, M.S.,
  Maharaj, T., Fischer, A., Courville, A., Bengio, Y., et~al.: A closer look at
  memorization in deep networks. In: International Conference on Machine
  Learning. pp. 233--242. PMLR (2017)

\bibitem{bao2013incoherent}
Bao, Y., Jiang, H., Dai, L., Liu, C.: Incoherent training of deep neural
  networks to de-correlate bottleneck features for speech recognition. In:
  International Conference on Acoustics, Speech and Signal Processing. pp.
  6980--6984 (2013)

\bibitem{barron1993universal}
Barron, A.R.: Universal approximation bounds for superpositions of a sigmoidal
  function. IEEE Transactions on Information theory pp. 930--945 (1993)

\bibitem{barron1994approximation}
Barron, A.R.: Approximation and estimation bounds for artificial neural
  networks. Machine Learning pp. 115--133 (1994)

\bibitem{bartlett2002rademacher}
Bartlett, P.L., Mendelson, S.: Rademacher and gaussian complexities: Risk
  bounds and structural results. Journal of Machine Learning Research pp.
  463--482 (2002)

\bibitem{bietti2019kernel}
Bietti, A., Mialon, G., Chen, D., Mairal, J.: A kernel perspective for
  regularizing deep neural networks. In: International Conference on Machine
  Learning. pp. 664--674 (2019)

\bibitem{bubeck2021universal}
Bubeck, S., Sellke, M.: A universal law of robustness via isoperimetry. Neural
  Information Processing Systems (Neurips)  (2021)

\bibitem{cogswell2015reducing}
Cogswell, M., Ahmed, F., Girshick, R.B., Zitnick, L., Batra, D.: Reducing
  overfitting in deep networks by decorrelating representations. In:
  International Conference on Learning Representations (2016)

\bibitem{deng2021discovering}
Deng, H., Ren, Q., Chen, X., Zhang, H., Ren, J., Zhang, Q.: Discovering and
  explaining the representation bottleneck of dnns. arXiv preprint
  arXiv:2111.06236  (2021)

\bibitem{deng2021adversarial}
Deng, Z., Zhang, L., Vodrahalli, K., Kawaguchi, K., Zou, J.: Adversarial
  training helps transfer learning via better representations. Neural
  Information Processing Systems (Neurips)  (2021)

\bibitem{du2020few}
Du, S.S., Hu, W., Kakade, S.M., Lee, J.D., Lei, Q.: Few-shot learning via
  learning the representation, provably. International Conference on Learning
  Representations  (2021)

\bibitem{dziugaite2017computing}
Dziugaite, G.K., Roy, D.M.: Computing nonvacuous generalization bounds for deep
  (stochastic) neural networks with many more parameters than training data.
  arXiv preprint arXiv:1703.11008  (2017)

\bibitem{foret2020sharpness}
Foret, P., Kleiner, A., Mobahi, H., Neyshabur, B.: Sharpness-aware minimization
  for efficiently improving generalization. arXiv preprint arXiv:2010.01412
  (2020)

\bibitem{golan2018deep}
Golan, I., El-Yaniv, R.: Deep anomaly detection using geometric
  transformations. In: Advances in Neural Information Processing Systems. pp.
  9758--9769 (2018)

\bibitem{golowich2018size}
Golowich, N., Rakhlin, A., Shamir, O.: Size-independent sample complexity of
  neural networks. In: Conference On Learning Theory. pp. 297--299 (2018)

\bibitem{goodfellow2016deep}
Goodfellow, I., Bengio, Y., Courville, A., Bengio, Y.: Deep learning. MIT Press
  (2016)

\bibitem{he2016deep}
He, K., Zhang, X., Ren, S., Sun, J.: Deep residual learning for image
  recognition. In: Proceedings of the IEEE conference on computer vision and
  pattern recognition. pp. 770--778 (2016)

\bibitem{hinton2012deep}
Hinton, G., Deng, L., Yu, D., Dahl, G.E., Mohamed, A.r., Jaitly, N., Senior,
  A., Vanhoucke, V., Nguyen, P., Sainath, T.N., et~al.: Deep neural networks
  for acoustic modeling in speech recognition: The shared views of four
  research groups. Signal processing magazine  \textbf{29}(6),  82--97 (2012)

\bibitem{hinton2012improving}
Hinton, G.E., Srivastava, N., Krizhevsky, A., Sutskever, I., Salakhutdinov,
  R.R.: Improving neural networks by preventing co-adaptation of feature
  detectors. arXiv preprint arXiv:1207.0580  (2012)

\bibitem{huang2017densely}
Huang, G., Liu, Z., Van Der~Maaten, L., Weinberger, K.Q.: Densely connected
  convolutional networks. In: Proceedings of the IEEE conference on computer
  vision and pattern recognition. pp. 4700--4708 (2017)

\bibitem{pmlr-v99-ji19a}
Ji, Z., Telgarsky, M.: The implicit bias of gradient descent on nonseparable
  data. In: Proceedings of the Thirty-Second Conference on Learning Theory. pp.
  1772--1798 (2019)

\bibitem{jiang2019fantastic}
Jiang, Y., Neyshabur, B., Mobahi, H., Krishnan, D., Bengio, S.: Fantastic
  generalization measures and where to find them. International Conference on
  Learning Representations  (2019)

\bibitem{kalimeris2019sgd}
Kalimeris, D., Kaplun, G., Nakkiran, P., Edelman, B., Yang, T., Barak, B.,
  Zhang, H.: Sgd on neural networks learns functions of increasing complexity.
  Neural Information Processing Systems  \textbf{32},  3496--3506 (2019)

\bibitem{kawaguchi2019depth}
Kawaguchi, K., Bengio, Y.: Depth with nonlinearity creates no bad local minima
  in resnets. Neural Networks  \textbf{118},  167--174 (2019)

\bibitem{kawaguchi2019gradient}
Kawaguchi, K., Huang, J.: Gradient descent finds global minima for
  generalizable deep neural networks of practical sizes. In: 2019 57th Annual
  Allerton Conference on Communication, Control, and Computing (Allerton). pp.
  92--99. IEEE (2019)

\bibitem{kawaguchi2017generalization}
Kawaguchi, K., Kaelbling, L.P., Bengio, Y.: Generalization in deep learning.
  arXiv preprint arXiv:1710.05468  (2017)

\bibitem{kornblith2021better}
Kornblith, S., Chen, T., Lee, H., Norouzi, M.: Why do better loss functions
  lead to less transferable features? Advances in Neural Information Processing
  Systems  \textbf{34} (2021)

\bibitem{krizhevsky2009learning}
Krizhevsky, A., Hinton, G., et~al.: Learning multiple layers of features from
  tiny images  (2009)

\bibitem{krizhevsky2012imagenet}
Krizhevsky, A., Sutskever, I., Hinton, G.E.: Imagenet classification with deep
  convolutional neural networks. In: Advances in Neural Information Processing
  Systems (2012)

\bibitem{kukavcka2017regularization}
Kuka{\v{c}}ka, J., Golkov, V., Cremers, D.: Regularization for deep learning: A
  taxonomy. arXiv preprint arXiv:1710.10686  (2017)

\bibitem{kwok2012priors}
Kwok, J.T., Adams, R.P.: Priors for diversity in generative latent variable
  models. In: Advances in Neural Information Processing Systems. pp. 2996--3004
  (2012)

\bibitem{laakom2022efficient}
Laakom, F., Raitoharju, J., Iosifidis, A., Gabbouj, M.: Efficient cnn with
  uncorrelated bag of features pooling. In: 2022 IEEE Symposium Series on
  Computational Intelligence (SSCI) (2022)

\bibitem{laakom2022reducing}
Laakom, F., Raitoharju, J., Iosifidis, A., Gabbouj, M.: Reducing redundancy in
  the bottleneck representation of the autoencoders. arXiv preprint
  arXiv:2202.04629  (2022)

\bibitem{laakom2023wld}
Laakom, F., Raitoharju, J., Iosifidis, A., Gabbouj, M.: Wld-reg: A
  data-dependent within-layer diversity regularizer. the 37th AAAI Conference
  on Artificial Intelligence  (2023)

\bibitem{lecun1998gradient}
LeCun, Y., Bottou, L., Bengio, Y., Haffner, P.: Gradient-based learning applied
  to document recognition. Proceedings of the IEEE  \textbf{86}(11),
  2278--2324 (1998)

\bibitem{lee2019meta}
Lee, H.B., Nam, T., Yang, E., Hwang, S.J.: Meta dropout: Learning to perturb
  latent features for generalization. In: International Conference on Learning
  Representations (2019)

\bibitem{li2016improved}
Li, Z., Gong, B., Yang, T.: Improved dropout for shallow and deep learning. In:
  Advances in Neural Information Processing Systems. pp. 2523--2531 (2016)

\bibitem{liao2019generalization}
Liao, Q., Miranda, B., Rosasco, L., Banburski, A., Liang, R., Hidary, J.,
  Poggio, T.: Generalization puzzles in deep networks. International Conference
  on Learning Representations  (2020)

\bibitem{mao2020multitask}
Mao, C., Gupta, A., Nitin, V., Ray, B., Song, S., Yang, J., Vondrick, C.:
  Multitask learning strengthens adversarial robustness. In: European
  Conference on Computer Vision. pp. 158--174. Springer (2020)

\bibitem{JMLR_ss}
Maurer, A., Pontil, M., Romera-Paredes, B.: The benefit of multitask
  representation learning. Journal of Machine Learning Research  (2016)

\bibitem{nagarajan2019uniform}
Nagarajan, V., Kolter, J.Z.: Uniform convergence may be unable to explain
  generalization in deep learning. In: Advances in Neural Information
  Processing Systems (2019)

\bibitem{neyshabur2017exploring}
Neyshabur, B., Bhojanapalli, S., McAllester, D., Srebro, N.: Exploring
  generalization in deep learning. NIPS  (2017)

\bibitem{neyshabur2018role}
Neyshabur, B., Li, Z., Bhojanapalli, S., LeCun, Y., Srebro, N.: The role of
  over-parametrization in generalization of neural networks. In: International
  Conference on Learning Representations (2018)

\bibitem{NEURIPS2019_36ab6265}
Pinot, R., Meunier, L., Araujo, A., Kashima, H., Yger, F., Gouy-Pailler, C.,
  Atif, J.: Theoretical evidence for adversarial robustness through
  randomization. In: Advances in Neural Information Processing Systems
  (Neurips) (2019)

\bibitem{poggio2017theory}
Poggio, T., Kawaguchi, K., Liao, Q., Miranda, B., Rosasco, L., Boix, X.,
  Hidary, J., Mhaskar, H.: Theory of deep learning {III}: explaining the
  non-overfitting puzzle. arXiv preprint arXiv:1801.00173  (2017)

\bibitem{rodriguez2021tighter}
Rodriguez-Galvez, B., Bassi, G., Thobaben, R., Skoglund, M.: Tighter expected
  generalization error bounds via wasserstein distance. Advances in Neural
  Information Processing Systems  (2021)

\bibitem{russakovsky2015imagenet}
Russakovsky, O., Deng, J., Su, H., Krause, J., Satheesh, S., Ma, S., Huang, Z.,
  Karpathy, A., Khosla, A., Bernstein, M., et~al.: Imagenet large scale visual
  recognition challenge. International journal of computer vision  (2015)

\bibitem{shalev2014understanding}
Shalev-Shwartz, S., Ben-David, S.: Understanding machine learning: From theory
  to algorithms. Cambridge university press (2014)

\bibitem{sokolic2016lessons}
Sokolic, J., Giryes, R., Sapiro, G., Rodrigues, M.R.: Lessons from the
  rademacher complexity for deep learning  (2016)

\bibitem{sontag1998vc}
Sontag, E.D.: {VC} dimension of neural networks. NATO ASI Series F Computer and
  Systems Sciences pp. 69--96 (1998)

\bibitem{valiant1984theory}
Valiant, L.: A theory ofthe learnable. Commun. ofthe ACM  \textbf{27}(1),
  134--1 (1984)

\bibitem{volhejn2020does}
Volhejn, V., Lampert, C.: Does sgd implicitly optimize for smoothness? In: DAGM
  German Conference on Pattern Recognition. pp. 246--259. Springer (2020)

\bibitem{wang2021towards}
Wang, X., Chen, X., Du, S.S., Tian, Y.: Towards demystifying representation
  learning with non-contrastive self-supervision. arXiv preprint
  arXiv:2110.04947  (2021)

\bibitem{wu2021wider}
Wu, B., Chen, J., Cai, D., He, X., Gu, Q.: Do wider neural networks really help
  adversarial robustness? Advances in Neural Information Processing Systems
  \textbf{34} (2021)

\bibitem{xie2017diverse}
Xie, B., Liang, Y., Song, L.: Diverse neural network learns true target
  functions. In: Artificial Intelligence and Statistics. pp. 1216--1224 (2017)

\bibitem{xie2015generalization}
Xie, P., Deng, Y., Xing, E.: On the generalization error bounds of neural
  networks under diversity-inducing mutual angular regularization. arXiv
  preprint arXiv:1511.07110  (2015)

\bibitem{xie2017uncorrelation}
Xie, P., Singh, A., Xing, E.P.: Uncorrelation and evenness: a new
  diversity-promoting regularizer. In: International Conference on Machine
  Learning. pp. 3811--3820 (2017)

\bibitem{xie2017aggregated}
Xie, S., Girshick, R., Doll{\'a}r, P., Tu, Z., He, K.: Aggregated residual
  transformations for deep neural networks. In: Proceedings of the IEEE
  conference on computer vision and pattern recognition. pp. 1492--1500 (2017)

\bibitem{yu2011diversity}
Yu, Y., Li, Y.F., Zhou, Z.H.: Diversity regularized machine. In: International
  Joint Conference on Artificial Intelligence (2011)

\bibitem{BMVC2016_87}
Zagoruyko, S., Komodakis, N.: Wide residual networks. In: Proceedings of the
  British Machine Vision Conference (BMVC) (2016)

\bibitem{zhai2018adaptive}
Zhai, K., Wang, H.: Adaptive dropout with rademacher complexity regularization.
  In: International Conference on Learning Representations (2018)

\bibitem{zhang2018mixup}
Zhang, H., Cisse, M., Dauphin, Y.N., Lopez-Paz, D.: mixup: Beyond empirical
  risk minimization. International Conference on Learning Representations 2018
  (2018)

\bibitem{zou2021benefits}
Zou, D., Wu, J., Gu, Q., Foster, D.P., Kakade, S., et~al.: The benefits of
  implicit regularization from sgd in least squares problems. Neural
  Information Processing Systems  \textbf{34} (2021)

\end{thebibliography}
%




\end{document}